**Title:**

Cyborg Insect Factory: Automatic Assembly System to Build up Insect-computer Hybrid Robot Based on Vision-guided Robotic Arm Manipulation of Custom Bipolar Electrodes

**Authors:**

Qifeng Lin[1], Nghia Vuong[1], Kewei Song[1], Phuoc Thanh Tran-Ngoc[1], Greg Angelo Gonzales Nonato[1], Hirotaka Sato[1*]

**Affiliations:**
[1]School of Mechanical & Aerospace Engineering, Nanyang Technological University; 50 Nanyang Avenue, 639798, Singapore.

*Corresponding author. Email: hirosato@ntu.edu.sg

**Abstract:**
The advancement of insect-computer hybrid robots holds significant promise for navigating complex terrains and enhancing robotics applications. This study introduced an automatic assembly method for insect-computer hybrid robots, which was accomplished by mounting backpack with precise implantation of custom-designed bipolar electrodes. We developed a stimulation protocol for the intersegmental membrane between pronotum and mesothorax of the Madagascar hissing cockroach, allowing for bipolar electrodes' automatic implantation using a robotic arm. The assembly process was integrated with a deep learning-based vision system to accurately identify the implantation site, and a dedicated structure to fix the insect (68 s for the whole assembly process). The automatically assembled hybrid robots demonstrated steering control (over 70 degrees for 0.4 s stimulation) and deceleration control (68.2% speed reduction for 0.4 s stimulation), matching the performance of manually assembled systems. Furthermore, a multi-agent system consisting of 4 hybrid robots successfully covered obstructed outdoor terrain (80.25% for 10 minutes 31 seconds), highlighting the feasibility of mass-producing these systems for practical applications. The proposed automatic assembly strategy reduced preparation time for the insect-computer hybrid robots while maintaining their precise control, laying a foundation for scalable production and deployment in real-world applications.

**One-Sentence Summary:** An automatic method for assembling insect-computer hybrids improves efficiency and scalability, enabling locomotion control and terrain coverage.

# 1. INTRODUCTION

Various insect-scale robots have been developed offering significant advantages in maneuverability within complex and narrow terrains (*1–5*). These benefits have driven the development both mechanically structured robots and insect-computer hybrid robots (i.e., biobots or cyborg insects). Despite their similar size, insect-computer hybrid robots provide distinct advantages in terms of self-locomotion energy sources (*3,6,7*) and adaptability to challenging terrains (*8*). Hence, more and more applications with insect-computer hybrid robots have been discussed to make full use of their potential as a robotic platform (*3,9–13*).

To achieve locomotion control of insects, stimulation electrodes targeting their muscles, neuron systems, and sensory organs have been tested (*7,14–17*). To facilitate the stimulation effect, electrodes, both invasive ones (*2,3,15*) and non-invasive ones (*7*), have been manually implanted into the target body parts of the insects. However, insects' tiny and delicate body structures made the manual surgery process time-consuming and difficult (*15*). Furthermore, the outcome of the surgery was highly influenced by the human's operation (*2,3,15*), which might lead to strict requirement to the operator's handy work and risk of insects' unnecessary injury. As the surgery skills of the operators could be different, even using the same implantation method, insect-computer hybrid robots produced by different operators could behave differently to some extent.

Therefore, to ensure consistent production of insect-computer hybrid robots, it is essential to transition from manual to automatic assembly processes. This is particularly important for applications requiring large numbers of these systems, such as post-disaster search and rescue or factory inspections, where multiple agents are more efficient than a single one (*18,19*). Achieving automatic assembly for insect-computer hybrid robots is thus a critical and urgent task to enable mass production. So far, Madagascar hissing cockroach has been used for different applications and proved as a powerful platform (*3,8,13,20*). Therefore, this insect was the focus of our study on mass production. Although the body structure of these cockroaches was generally similar, individual size variations made uniform localization of the implantation site challenging, unlike in some assembly tasks for specific shapes of mechanical parts (*21*). Consequently, the precise localization of the implantation site must be determined using deep learning techniques.

To control the locomotion direction of Madagascar hissing cockroach, researchers have developed stimulation protocols on the insects' antennae and abdomen (*3,15,22*). The antennae were critical sensory organ for detecting and navigating around the obstacles (*23*). Cockroaches' locomotion was found to be highly influenced by the tactile stimuli (*24*). Subsequent tests showed that electrical stimulation of the antennae effectively induced directional turning in the insects (*7,15*). However, the antennae are soft, fragile, and tiny (*15*) (diameter was 0.6-0.7 mm, Fig. 1B), making them difficult to fix and implant electrodes inside. Moreover, antennae could be used for the hybrid robots to deal with the obstacles naturally themselves (*7,25*). Implantation on the antennae could destroy the insects' nature to negotiate with the obstacles (*7,26*). Hence, antennae were not considered to the target stimulation site in this study.

Stimulation on side of abdomen was also proved effectiveness to influence the locomotion direction of the insects (*3*). Nevertheless, abdominal cuticles of cockroach were short and thin (third abdominal cuticle, 3.8-5.0 mm in length, 0.2-0.3 mm in thickness), which were hard to fix automatically for the electrode implantation. As a result, a new stimulation site should be considered. Inspired by studies on similar terrestrial platforms like Zophobas morio, which altered its locomotion direction when stimulated on the pronotum and elytra (*2*), we hypothesized that stimulation of the pronotum could similarly alter the cockroach's direction. And compared to the abdominal cuticles, the pronotum cuticle was larger and thicker (11.6-13.4 mm in length, 0.5-0.6 mm in thickness), making it easier to fix. Additionally, similar to the interspace between abdominal segments (*7*), an intersegmental membrane existed between the pronotum and the mesothorax, which could serve as a target stimulation site for directional control of the insect (Fig. 1C).

Our work in this paper proposed a new stimulation protocol to direct the insect-computer hybrid robot. Insects' neural activities, forelegs movement, and their locomotion change during the electrical stimulation were recorded to study their reaction against the electrical stimulation. A backpack with the integration of microcontroller, stimulation electrodes, and mounting device was developed for the insect (Fig. 1B, Fig. 1C). The hybrid robot could be wirelessly controlled to steer and stop. Afterwards, automatic assembly using a slide motor with a structure for insect's fixation, intel RealSense D435 camera, Robotiq Hand-e gripper and Universal Robot UR3e (Fig. 1A) was designed based on the visual detection of the target body position. The insect was firstly fixed in a structure which was mounted on a slider and driven by the motor. After the backpack being assembled to the insect, the structure to fix insect was released. The assembly process for the hybrid robot finished. 5 hybrid robots assembled automatically were studied on their locomotion control, including steering, and deceleration stimulation to compare with the manually assembled ones. A team of 4 hybrid robots was demonstrated to cover an outdoor uneven terrain with UWB localization system.

## 2. RESULTS AND DISCUSSION
### 2.1 Bipolar Electrode

Due to the difficulty of fixing the antennae and abdomen and implanting electrodes using the robotic arm, this study focused solely on the pronotum. To induce the cockroach's turning locomotion and minimize implantation sites, a pair of bipolar electrodes were designed and implanted on the left and right sides of the intersegmental membrane between the pronotum and mesothorax (Fig. 1C, and Fig. 2A). Each bipolar electrode consists of a copper pattern for electrical signal transmission and is designed with a microneedle structure for rapid puncture of the intersegmental membrane, along with a hook to prevent detachment after implantation.

Given the complex structural features and multi-material composition of the bipolar electrodes (including both plastic and metal components), specialized fabrication processes were required. The combination of multi-material 3D printing technology and electroless plating process provided an effective means for 3D electronic structures with this type of spacing structure function and electrical signal carrying function *(27–29)* (Fig. 2B). Initially, multi-material DLP3D printing was used to create the precursor structure of the bipolar electrodes, composed of a normal resin and an active precursor. The active precursor contains a catalytic factor that facilitates selective metal deposition during electroless plating process, resulting in metallization on both sides of the structure. Fig. 2C shows the precursor and the final bipolar electrode with selectively deposited copper.

To ensure effective implantation and electrical stimulation, the bipolar electrodes were anticipated to exhibit high hardness and toughness. Consequently, we selected ABS-like photosensitive resins (*30*) as the material for electrode fabrication (whether using normal resin or active precursor). The bipolar electrode showed smooth stress distribution and controllable deformation during implantation, without damage or yielding (finite element simulations, Fig. 2D, i-iv). And the designed bipolar electrodes could be securely implanted by monitoring the stabbing stress analysis of the microtip at the membrane (Fig. 2D, v).

To address potential issues with metal plating, which could be severed or delaminated during implantation, we calibrated the plating adhesion according to the ASTM D3359-09 standard. We introduced a chemical etching process in the selective electroless plating to enhance the adhesion, achieving a high grade of 4B (Fig. 2E). Additionally, the implanted part within the cockroach, composed of soft tissue (intersegmental membrane) and electrolyte solution, exerted minimal cutting force on the electrode, thereby reducing impact (Fig. 2A). Fig. 2F shows the impedance profile of a plated layer on the bipolar electrode. The impedance of the bipolar electrode was stably below 70 Ω, significantly lower than previously demonstrated non-invasive electrodes (over 1000 Ω) (*7*). This low impedance could enable more intense stimulation and insect's more obvious reaction.

The conductivity of the plating was calculated to be $3.12 \times 10^7$ S/m, while the plating thickness was 2.5 um. Due to the copper-replacing-nickel plating method, the electrodes achieved selective metallization of copper. After 5 minutes of immersion in the plating solution, the plating thickness began to increase significantly (Fig. 2G). A thicker plating layer enhanced conductivity, reduced impedance, and improved corrosion resistance, but it might also increase parasitic capacitance. In this study, a plating time of 16 minutes was selected to optimize plating thickness and conductivity. By controlling the duration of the electroless plating process, the thickness of the plated layer on the bipolar electrodes was precisely modulated, allowing for the fine-tuning of conductivity and other electrochemical properties.

**2.2 Stimulation Protocol on Insect's Pronotum**

To determine an effective stimulation voltage, neural activities in the insects' neck region were recorded and analyzed (Fig. 3A). The number of detected neural spikes increased gradually from 0.5 V to 3.0 V and plateaued between 3.0 V and 3.5 V. However, when the stimulation voltage was increased to 4.0 V, the average number of spikes decreased by 23.5%. The initial increase in neural spikes with the first few voltages suggests that the insects responded more intensely to higher stimulation. However, the plateau observed beyond 3.0 V indicates that electrical stimulation above 3.0 V could not elicit a stronger response from the insects. The reduction in neural activity at 4.0 V may be due to damage to the insects' neural system. To avoid unnecessary damage to the insects and to ensure an effective stimulation, a voltage of 3.0 V was considered optimal for the subsequent discussions.

Since the pronotum of the insect is directly connected to its forelegs (Fig. 3B) and the forelegs direct its locomotion (*31*), it is important to examine the forelegs' status during stimulation. When stimulation happened on one side of the pronotum, the foreleg on that side exhibited contraction (Fig. 3B, movie S1). The foreleg remained in a contracted state until the stimulation ceased (movie S1). These observations confirm that the forelegs were affected by the electrical stimulation. As the forelegs directs the insects in walking (*31*), the forelegs' reaction to the stimulation further suggests their usage to controlling insect's orientation. Consequently, a locomotion study was undertaken in the following work.

To simplify future automatic assembly process, the stimulation bipolar electrodes, microcontroller, and mounting parts were first assembled as a backpack. Backpacks were assembled with cockroaches manually to study the locomotion control and then were assembled automatically to have a fair comparison. Manually assembled hybrid robots were tested for the locomotion control (N = 5 insects). Based on the cockroaches' reaction against the electrical stimulation on the two sides of pronotum (Fig. 3C), the stimulation electrodes implanted to the cockroach's intersegmental membrane between pronotum and mesothorax were able to turn the insects. Specifically, the electrical stimulation caused the cockroaches to turn with average angles of 68.0 degrees for left turns and 82.6 degrees for right turns. The maximum angular speeds achieved 275.8 degree/s for left turns and 298.2 degree/s for right turns. Additionally, combined with the contraction of the forelegs during the stimulation (Fig. 3B), it indicated that the implantation on the cockroach's intersegmental membrane between pronotum and mesothorax successfully stimulated the forelegs and the cockroach could be steered if one of its forelegs was stimulated.

Our new stimulation protocol demonstrates a substantial enhancement in performance compared to previous non-invasive electrode-based methods (*7*). Specifically, it increases the maximum steering speed by over five times and improves the turning angle by more than 76.6%. Additionally, this protocol achieves these results using only 40% of the stimulation time and 75% of the stimulation voltage. These improvements suggest that our protocol not only induces more intense turning responses but does so more efficiently, both in terms of time and energy consumption. The reduction in required stimulation time and voltage signifies a more resource-efficient approach, which can enhance the overall performance of the hybrid robot. By achieving greater control with reduced energy expenditure, this new protocol offers significant advantages in operational efficiency and cost-effectiveness. It represents a notable advancement in optimizing locomotion control for insect-computer hybrid robots, making it a valuable tool for practical applications where quick, and energy-efficient control is essential.

Besides, we have found the cockroach decelerated when the electrical stimulation was outputted from outer electrodes of the two bipolar electrodes (Fig. 2A). This achievement marks the first successful implementation of deceleration control in the field of insect-computer hybrid robots. During the real application, deceleration could help the insect-computer hybrid robot to avoid the collision after detecting its front obstacles (*32*). Following 0.33 seconds of stimulation, the average walking speed of the cockroaches decreased from 6.2 cm/s to a minimum of 1.5 cm/s (Fig. 3C, iii). This reduction corresponds to a decline from 112.9% to 27.6% of their body length per second (with an average body length of 5.5 cm), indicating an 85.3% deceleration relative to their body length. The standard deviation of the minimum speed during stimulation was 1.3 cm/s, which is half of the standard deviation of the pre-stimulation walking speed (2.6 cm/s). This smaller standard deviation in the minimum decelerated speed highlights the consistency of the deceleration stimulation. The simultaneous contraction of both forelegs during stimulation (Fig. 3B, iv) further indicates that the deceleration trend in cockroaches was directly associated with the stimulation of their forelegs.

Based on the above, with the new stimulation protocol on the pronotum, both steering and deceleration control of the insect-computer hybrid robot realized.

**2.3 Vision-based Automatic Assembly of Insect-computer Hybrid Robots**

Automatic assembly of hybrid robots included the following steps: 1) fixing the pronotum and mesothorax of an anesthetized cockroach and exposing the intersegmental membrane; 2) locating the reference point for electrode implantation; 3) grasping a backpack with the gripper on the robotic arm; 4) implanting the bipolar electrodes into the exposed membrane with the robotic arm; 5) pressing the backpack until its mounting branches hooks the insect's metathorax; 6) releasing the backpack from the gripper; and 7) retracting the structure to free the insect (Fig. 4A).

For successful assembly, certain key considerations must be addressed. First, the insect's pronotum and mesothorax must be fixed to expose their intersegmental membrane (Fig. 4A, ii). Next, precise identification of the reference point on the pronotum is crucial for the accurate implantation of the bipolar electrodes by the robotic arm. Finally, the robotic arm should be manipulated to assemble the backpack with the insect at the appropriate angle for secure assembly.

**2.3.1 Exposure of Intersegmental Membrane between Pronotum and Mesothorax**

The intersegmental membrane between the pronotum and mesothorax is covered by the hard cuticle of the connected pronotum (Fig. 4B, ii). To implant bipolar electrodes within this membrane, the posterior pronotum must be firstly lifted from the mesothorax (Fig. 4B, ii). A structure with Rod A and Rod B (Fig. 4B, iii), was designed for this purpose. Rod A applied pressure to the anterior pronotum, while Rod B pressed on the mesothorax, thereby exposing the membrane for electrodes implantation (Fig. 4B, iii).

Initially, Rod A was positioned 4.0 mm above the platform, which corresponded to a lowered distance, d, of 0 mm. The relationship between the lowered distance, d, and the lifting height, h, of the pronotum is shown in Fig. 4B, iii, demonstrating the progressive increase in intersegmental membrane exposure as the structure is lowered. The bipolar electrode, with a thickness of 0.6 mm, requires that the lifting height, h, consistently exceeds this value to provide sufficient space for implantation. On average, the height, h, reaches 1.9 mm when d is set at 3.5 mm or higher, which provides adequate exposure for implantation.

Further increasing d beyond 3.5 mm causes the lifting height, h, to stabilize around 1.9 mm, indicating that additional lowering of the structure does not significantly increase membrane exposure (Student's t-test for d = 3.5 and 4.0 mm, $P$ = 0.31). Since all lifting heights at both d = 3.5 mm and d = 4.0 mm exceed more than twice the thickness of the bipolar electrodes, d = 3.5 mm was chosen to avoid unnecessary pressure on the insect body while ensuring sufficient membrane exposure for bipolar electrodes insertion. This approach minimizes the potential injury to the insect, ensuring its physical integrity throughout the implantation process.

**2.3.2 Detection of Pronotum Using Deep Learning**

To implant the bipolar electrodes into the exposed intersegmental membrane from section 2.3.1, computer vision was used to locate the position of the membrane. Since bipolar electrodes on both sides of the backpack needed to be implanted simultaneously and symmetrically on the insect's intersegmental membrane, the middle point of the posterior pronotum edge was set as the reference point, $p_R$, for the implantation process. However, the pronotum has variations in size and shape across each cockroach (Fig. S1). Moreover, even though the cockroaches' pronotum and mesothorax were restricted by the structure to fix, their pronotum position may still vary in x- and y- directions (Fig. 4B, iii, Fig. S1). Hence, a deep learning-based computer vision model should be used to identify the pronotum and subsequently, the position of the reference point on the pronotum, $p_R$.

Several segmentation models, including UNet, Deeplabv3, TransUNet, and Segment Anything, were evaluated for their accuracy in pronotum segmentation. Performance metrics were based on mean intersection over union (mIoU) score, mean Dice similarity coefficient (mDSC), and mean squared error (MSE) of $p_R$ prediction, as shown in Table 1.

TransUNet model outperformed the other models in terms of mIoU, mDSC and MSE metrics. This may be attributed to its hybrid encoder, leveraging the advantages of the Transformer architecture while still maintaining locality through its CNN counterpart. Deeplabv3 had similar performance to TransUNet in the mIoU and mDSC metrics but was significantly worse in the MSE metric as it was generally able to segment out most of the pronotum but did not precisely identify the lower border of the pronotum, causing the $p_R$ predictions to be further than the ground truth. Subsequently, DSC loss and Boundary Difference Over Union (bDoU) loss (*33*) were compared with BCE loss to achieve improved boundary results. The results of the loss function evaluation were shown in Table 2 in terms of mIoU score, mDSC score, and MSE of $p_R$ prediction. The DSC and bDoU loss functions generally outperformed the BCE loss function, as they evaluated the spatial overlap of the prediction and ground truth masks, so the background pixels were generally ignored when calculating the loss. The DSC loss was superior to the bDoU loss, as it captured features from the entire pronotum, whereas the latter focused primarily on the pronotum's border. Hence, the chosen deep learning solution was the TransUNet model trained with the DSC loss function.

**2.3.3 Manipulation of Robotic Arm**

To assemble the backpack on the insect, the robotic arm first used the camera to scan the insect and identify the reference point for bipolar electrode implantation (Section 2.3.2). To ensure a secure grasp on the backpack and stable electrode implantation, the Robotiq Hand-e gripper was selected, offering a gripping force of up to 185 N. Considering the combined weight of the Robotiq Hand-e (1.0 kg), RealSense D435 camera (75.0 g), and the backpack (2.3 g), the Universal Robots UR3e was chosen as the robotic arm, given its 3.0 kg payload capacity and 0.03 mm pose repeatability at full payload (*34*). Additionally, the UR3e's 500 mm reach accommodates the RealSense D435's minimum depth sensing distance at maximum resolution (~280.0 mm).

To accurately detect the reference point for bipolar electrode implantation, the robotic arm was positioned vertically, with its gripper and camera oriented directly downward. After the camera confirming the position of reference point, robotic arm went down to the position of backpack. The backpack was always placed to the backpack holder (Fig. 1A). Hence, waypoints for robotic arm to grasp the backpack were consistent. Once grasping the backpack, the robotic arm carried the backpack to the same pre-implant waypoint. Afterwards, the robotic arm carried the backpack and implanted the bipolar electrodes inside the exposed intersegmental membrane. During this process, as the manipulation space was limited for robotic arm ($6.5 \times 3.5 \times 2.5$ cm$^3$, between the insect and the structure to fix it), an optimal angle should be determined to facilitate implantation of the bipolar electrodes without any collision. Since the insect's pronotum is symmetrical and it is positioned against the marked spot (Fig. 4B, iii), only the pitch angle ($\alpha$, rotation around y-axis) needs to be considered, while roll and yaw angles can be disregarded.

During bipolar electrode implantation, collisions can occur in two scenarios: between the backpack and the 3D-designed structure, or between the backpack branches and the insect's dorsal cuticles. For both scenarios, we identified the pitch angles at which the backpack contacts the 3D structure ($\alpha_L$) and the insect's dorsal cuticles ($\alpha_U$), utilizing a sample of five insects. Our measurements (Fig. 4C, ii) revealed $\alpha_L$ to be $157.8 \pm 1.5$ degrees and $\alpha_U$ to be $167.5 \pm 2.2$ degrees (mean ± standard deviation). To minimize the risk of contact with either the insect or the fixation structure during implantation, the mid-point angle of 162.7 degrees was selected, ensuring both the accuracy and safety of the procedure (Movie S2).

After implanting the bipolar electrodes, the robotic arm pressed the backpack downward to hook the metathorax cuticle with the backpack's four branches, completing the assembly of the hybrid robot. The robotic arm then released the backpack and returned to its initial position to capture an image of the next fixed insect. Finally, the structure to fix the insect was retracted, allowing the next insect to be positioned and fixed on the platform. The entire assembly process took 68 seconds, demonstrating the efficiency of the proposed automated assembly approach for mass production (Movie S2).

## 2.4 Locomotion Control and Dispersion of Automatically Assembled Insect-computer Hybrid Robots

To evaluate the effectiveness of the automatic assembly, the previously established steering and stopping protocols were tested on five automatically assembled hybrid robots. The results show that the performance of these systems closely matched that of manually assembled ones, with a maximum steering speed of 240.0 degree/s for left turns and 273.5 degree/s for right turns, differing by less than 13% compared to the hand-assembled group (Fig. 5A, i). The average turning angles (Fig. 5A, ii) were 70.9 degrees for left and 79.5 degrees for right, with no significant differences observed (Student's t-test, $P = 0.62$ for left turns, $P = 0.50$ for right turns). This indicates that the automatic assembly maintained consistent steering control. The difference in average turning angles between left and right was reduced to 10.8%, which is a 50.2% reduction compared to manually assembled systems (21.7%). This suggests that the automatic assembly may contribute to more balanced directional control. The reason of this reduction could be that the automatic assembly avoided manual errors from the assembly operator.

For deceleration, the average walking speed decreased from 6.3 cm/s to 2.0 cm/s, consistent with the results from hand-assembled systems (Student's t-test between automatic assembled and manual assembled hybrid robots: $P = 0.21$). The similarity in deceleration performance further demonstrates that the automatic process does not compromise control quality (Fig. 3C, iii, Fig. 5A, iii).

These findings confirm that the automatically assembled insect-computer hybrid robots achieve comparable locomotion control to manually assembled systems, validating the effectiveness and precision of the automated assembly process.

With the benefit of automatic assembly, four insect-computer hybrid robots were assembled within 7 minutes 48 seconds). Compared to the previous time-consuming manual preparation (more than 1 hour for assembling one hybrid robot (*8*)), the automatic assembly significantly reduced the preparation time. Therefore, they could be applied timely based on the needs of the missions. For the tasks using multiple agents, terrain covering could be the most fundamental one (*8*). Hence, covering of an unknown obstructed outdoor terrain was demonstrated in this paper to strengthen the necessity and benefits of the mass production of the hybrid robots.

A team of 4 hybrid robots automatically prepared was released to an obstructed outdoor terrain ($2 \times 2$ m$^2$) with obstacles randomly placed inside (Fig. 5B, i). A UWB system was used to track the hybrid robots (Fig. 5B, ii). Each hybrid robot carried a UWB label to enable their own localization. They were boosted with methyl salicylate (*8*) before the release and then randomly stimulated every 10 s. The trajectory of each hybrid robot was tracked (Fig. 5B, iii). The combined coverage of all four insects increases steadily and reaches close to 80.25% after 10 minutes and 31 seconds (Fig. 5B, iv), demonstrating that deploying multiple hybrid robots significantly improves the overall coverage compared to any single insect alone. Comparing the single insect's coverage (from 14.00% to 45.75%), the whole team could achieve higher coverage (80.25%, 50.87 cm$^2$/s on average, Fig. 5B, iv). This covering performance showed the efficiency of the simple coverage strategy using the multiple hybrid robots. Even though some demonstrations were achieved by multiple agents (*8,13,35*), our covering mission based on the insect-computer hybrid robots were firstly achieved the covering of the outdoor obstructed terrain.

**3. CONCLUSION**
   This study developed an automatic assembly strategy for insect-computer hybrid robot based on a newly developed pronotum stimulation protocol. Locomotion studies and neural recordings on the hybrid robots reflected the reliability of the implantation and stimulation. Under this assembly procedure, the hybrid robots could be prepared in a short time, i.e., 68 s. With the benefit of the fast preparation, mass production of the hybrid robots could be realistic. A team of 4 hybrid robots was demonstrated to cover a terrain with a simple navigation algorithm. The covering result indicated the practical application of the swarm of the hybrid robots and meaningfulness of their mass production. In the future, factories for insect-computer hybrid robot could be built to satisfy the needs for fast preparation and application of the hybrid robots. Different sensors could be added to the backpack to develop applications on the inspection and search missions based on the requirements.

## 4. MATERIALS AND METHODS

### 4.1 Insect Platform

Male Madagascar hissing cockroaches (5-6 cm) were used for this study. These insects were commonly used for the hybrid robots (*7,15*), which had mature stimulation protocol to compare with our new one. All the cockroaches used were supplied carrots and water weekly. The cockroaches used for hybrid robot assembly were given 10 minutes to anesthetize under $CO_2$ and their backpacks were removed after the experiment finished.

### 4.2 Backpack

Backpack included three parts, bipolar electrodes, mounting structure and microcontroller. Below is the detailed information for these three parts.

### 4.2.1 Bipolar Electrodes Preparation

**a. Preparation of 3D printing ink**

ABS-like photosensitive polymer raw material (SeedTech Electronics Co., LTD) used in this study is an acrylonitrile butadiene styrene (ABS)–like polymer, which contains a light source initiator sensitive to 405 nm ultraviolet wavelength. The basic mechanical properties of the materials are shown in Table S1. This polymer exhibits strong mechanical properties after curing, following cross-linking and molding.

First, dissolve 15.4 g $NH_4Cl$ in 50 ml deionized water, then add 270 mg $PdCl_2$, stir and dissolve to obtain 50 ml saturated activation solution containing $Pd^{2+}$ 0.2 wt %. After the solution stood for 30 minutes, 12 ml of the upper clear part was added dropwisely to 38 ml ABS-like photosensitive resin with a 1000 RPM magnetic stirrer. Finally, the ink was stirred at 1200 RPM for 30 minutes to obtain 50 ml active precursor ($Pd^{2+}$ concentration is about 0.058 wt %).

**b. Bipolar electrodes' fabrication with multi-material 3D printing and electroless plating**

Bipolar electrodes were fabricated through multi-material DLP 3D printing and selective electroless plating (*27–29*). The multi-material DLP 3D printing platform uses a 6.1-inch 405 nm-90 W ultraviolet parallel light source with a light intensity of 89 MW/cm$^2$ and a light uniformity of 99%. During the printing process, we set different slice thicknesses and single-layer exposure times for different materials (normal resin and active precursor), as shown in Table S2.

Through the multi-material DLP 3D printing process, a composite structure of normal resin and active precursor was obtained, where the topology of the active precursor could be selectively deposited with metal (Cu) in the subsequent electroless plating process. Ni substitution was used to obtain the copper coating. Firstly, electroless Ni plating was performed and then selective electroless copper plating was achieved through metal substitution reaction. The nickel-plating bath (pH = 9, 70 °C) used in this process consisted of $NiSO_4·6H_2O$ and $NaH_2PO_2·H_2O$. For each printed multi-material part, the active precursor distributed on the resin substrate with the designed 3D topology throughout the bipolar electrode. After the printed part was immersed in the bath, the surface-exposed $Pd^{2+}$ ions were initially reduced to Pd monomers by the reducing agent (sodium hypophosphite monohydrate) in the plating bath, which acted as catalytically active metal cores to initiate the ELP reaction in specific microscopic areas, thereby achieving the targeted Ni metal deposition. Finally, for copper plating (i.e., Ni substitution), the plating bath used (pH = 12.2, 70 °C) consisted of $CuSO_4·5H_2O$ and HCHO. The active Pd monomer was also effective for electroless copper plating.

**c. Finite element analysis of 3D electrodes**

Finite element calculations were performed using Ansys by creating a 3D electrode model, assigning materials, meshing, and setting boundary conditions. Finite element analysis was performed to obtain the relationship between electrode implantation stress and implantation force, and to evaluate electrode damage during the process. The policy parameters are shown in Table 3.

The finite element analysis for membrane damage simulation was carried out using Lsdyna part in Ansys 19.0. In the mesh setup, a 2 mm mesh was set for the membrane part (Fig. 2A) to improve the computation time and a 1mm mesh was set for the microneedle structure of the bipolar electrodes (Fig. 2A) to ensure the accuracy of the simulation results. In the boundary conditions, membrane (i.e., implanted structure) was set as fixation and then a z-direction displacement of 50mm was set for the microneedle structure to achieve the implantation, after completing the setup the results of the simulation were calculated. The policy parameters are shown in Table 3.

### 4.2.2 Mounting Structure

Mounting structure was sticked with the microcontroller and bipolar electrodes. An inclined plane was designed for convenience of robotic arm's grasping and alignment hole was used to always place the backpack at the same position of backpack holder (Fig. 1A). There were also four mounting branches to hook the cockroach's metathorax (two branches for each side of metathorax, Fig. 1C).

### 4.2.3 Microcontroller

The microcontroller used to control the cockroach locomotion could communicate with a workstation via Sub-1 GHz. After receiving stimulation commands from the workstation, the microcontroller outputted the electrical signals through its 4 stimulation channels (Fig. 1D). Before the assembly for the hybrid robot, the microcontroller was sticked to the mounting structure with double side tapes. A lithium battery (3.7 V, 50 mAh) was used to power the microcontroller after the hybrid robot was assembled and before their locomotion control experiment started.

**4.3 Automatic Assembly of Insect-computer Hybrid Robots**

**4.3.1 Structure to Fix Insect and to Expose Intersegmental Membrane**

To expose the cockroach's intersegmental membrane between its pronotum and mesothorax, its pronotum should be lifted from the mesothorax. Hence, a structure with two rods (Rod A and Rod B, Fig. 4B, iii) should be developed. Rod B could press the cockroach's mesothorax and Rod A pressed pronotum' anterior part to lever its posterior part. Ten different cockroaches were tested with different fixation lowered distance to evaluate the suitable pronotum lifting height, h (Fig. 4B, iv).

The middle part of the designed 3D structure was skeletonized to facilitate the camera's subsequent detection of the electrode implantation point. The skeletonized part was a rectangle with an area of $66 \times 34$ mm$^2$ (Fig. 4B, i). With the skeletonized structure, the camera could successfully capture the image with cockroaches' intact pronotums (Fig. 4A, iii).

An anesthetized cockroach was placed to the platform with pronotum aligned against a marked spot (Fig. 4B, iii). This marked spot was designed 2 mm ahead of the Rod A (Fig. 4B, iii) to ensure that the cockroach's protruding cuticle on the anterior pronotum could be securely fixed by the Rod A. Subsequently, a slider motor drove the structure downwards to execute fixation. At this point, the cockroach's intersegmental membrane between its pronotum and mesothorax was fully exposed for the robotic arm's implantation of the bipolar electrodes. After the robotic arm completed the assembly of the anesthetized cockroach and the backpack, the structure to fix insect was retracted (Fig. 4A, viii).

**4.3.2 Identification of Implantation Reference Point Based on Deep Learning**

Generally deep learning models consists of Convolutional Neural Networks (CNNs) and more recently, Vision Transformers (ViTs). The latter recently has had growing interest from researchers for the field of computer vision, outperforming their CNN-based counterparts in several vision applications such as object classification and detection (*36*). However, ViTs require large amounts of data for training, such as the recent Segment Anything segmentation model which was trained on the SA-1B dataset containing over 1 billion masks and 11 million images (*37*). ViTs were found to lack locality inductive biases (relation of image pixels based on their position) as their self-attention layers use a global context, compared to CNNs which preserve locality information as convolution layers pass through the images like sliding windows (*38,39*). Hence, CNN-based models were mainly evaluated for our application, due to the small size of available cockroach data for training. The following models were trained and evaluated on our dataset of cockroaches: UNet (*40*), TransUNet (*41*), Deeplabv3 (*42*) and Segment Anything (*37*).

The UNet and Deeplabv3 were CNN-based models, with the Deeplabv3 model using the ResNet-101 (*43*), which was pretrained on ImageNet (*44*), as it's backbone. The TransUNet model was a hybrid model, having a CNN-Transformer hybrid encoder, combining ResNet-50 (*43*) and ViT-B (*38*), which was pretrained on ImageNet (*44*). Segment Anything was a promptable ViT-H (*38*) model and box prompts were used to specify the pronotum as the segmentation target. Apart from Segment Anything, the other models were trained on the cockroach dataset, with a batch size of 32, Adaptive

Moment Estimation as the optimizer and Binary Cross Entropy (BCE) as the loss function for 150 epochs.

A set of 29 unique cockroaches were fixed by the designed 3D structure. Meanwhile, the robot arm was placed at a consistent position to take their images with the Intel RealSense D435 camera. As the position of the 3D structure and camera were consistent for all samples, a 256 × 256-pixel crop of the original image, centered around the pronotum, was used as the training images for the models, to shorten the training and inference times. 20 of these images were used as test cases for model evaluation, while the others were used for training. Data augmentation was applied extensively to the training data to improve the robustness of the models against variances in the rotation and shape of pronotums. Asymmetrical scaling in the x- and y- axes was applied to the training samples using bilinear interpolation to generate more unique pronotum shapes to simulate the varying pronotum sizes of cockroaches. Subsequently, more training samples were generated by applying rotations between -30° and 30° to accommodate inconsistencies in the cockroaches pose when they are mounted on the 3D structure.

### 4.3.3 Automatic Manipulation of Robotic Arm

Once the cockroach was fixed and its intersegmental membrane was uncovered, intel RealSense D435 camera which mounted on the Robotiq Hand e gripper would take a picture of the cockroach through the skeletonized structure of the 3D designed structure (Fig. 4A, iii, Fig. 4B, i). Based on the model trained before, a reference point on the middle posterior edge of the pronotum could be found (green point on the Fig. 4A, iii). As intel RealSense D435 camera could also give the depth information, after hand-eye calibration with the robotic arm UR3e, the x, y, z position of the reference point relative to the base of the robotic arm could be identified. Then, robotic arm grasped the backpack, which was always placed on the backpack holder (Fig. 1A).

The bipolar electrodes of the backpack should be implanted inside the exposed intersegmental membrane (Fig. 4A, v). To achieve this motion, the robotic arm must be programmed to avoid collisions with both the cockroach and the structure to fix the cockroach. Given that the implantation reference position, denoted as $p_R$, was already determined, the implantation angle became a critical parameter. Since the cockroach was aligned with the predetermined marked spot and its pronotum was symmetrical, the implantation motion was simplified to consider only the pitch angle. Robotic arm grasped the backpack and placed the tips of bipolar electrodes under the cockroach's pronotum with different pitch angles until backpack touched the 3D structure (lower threshold, $α_L$) or the cockroach (upper threshold, $α_U$). N = 5 cockroaches were tested to collect the data for these two thresholds.

With the implantation reference point and implantation pitch angle, the bipolar electrodes could be implanted inside the insect's intersegmental membrane. Then, the backpack was pressed down until its mounting branches hooked cockroach's metathorax (Fig. 4A, vi). Finally, gripper released the backpack (Fig. 4A, vii) and robotic arm returned to the initial state to support the camera to take the picture of the next cockroach.

### 4.4 Neural Recording During the Stimulation

To check the stimulation reaction and to find appropriate stimulation strength, insects' neural reactions during the electrical stimulation period were recorded. Three cockroaches were anesthetized with $CO_2$ for 10 minutes and then dissected to expose their ventral nerve cords inside their neck. The designed bipolar electrode was implanted to the intersegmental membrane between pronotum and mesothorax to transfer the electrical stimulation. Microcontroller of the backpack was also used here to generate electrical stimulation. The nerve codes were rinsed with cockroach saline to secure that they could be found under microscope. Two probes were attached to a nerve code for recording the nerve signals passed. A ground pin was implanted to the cockroach's abdomen.

A single bipolar square-wave pulse (1 Hz, 1.0 s) was outputted from the microcontroller to the cockroach. The amplitude of the stimulation changed from 0.5 V to 4.0 V. Each type of stimulation was repeated three times. During the electrical stimulation, neural reaction recorded could contain some artefact signals. Hence, 50 ms neural signals started from 0 s, 0.5 s, and 1.0 s, i.e., the edge of pulse, were replaced with zero. After that, neural signals were filtered by a second-order Butterworth filter (300-5000 Hz). Neural spikes were detected with a threshold $T$.

$$T = 5 \times median\ (|x|/0.6745)$$

Where $x$ was the filtered signals. The detected neural spikes were marked with blue circles (Fig. 3A, ii) and counted at different stimulation voltages (Fig. 3A, iii)

### 4.5 Locomotion Control of Insect-computer Hybrid Robots

Five insect-computer hybrid robots (manually assembled and automatically assembled) were tested for locomotion control. Electrical stimulation with bipolar pulse wave (0.4 s, 3.0 V, 42 Hz) was used to stimulate the insects. For each stimulation type (turning right/left, deceleration), insects were stimulated five times. Their locomotion reaction was recorded with motion tracking system (VICON) for the data analysis.

### 4.6 Dispersion of Multiple Insect-computer Hybrid Robots

Four insect-computer hybrid robots were given 4 hours to rest after the assembly. They were then used for the covering mission on the obstructed terrain (2.0 × 2.0 m$^2$, Fig. 5B, i). Four UWB anchors were placed on the corners of a 3.6 × 3.6 m$^2$ (Fig. 5B, ii) to track the hybrid robots sticked with UWB labels. Hybrid robots were released from a corner of the target terrain with chemical booster, methyl salicylate (8) and then randomly stimulated (steered or decelerated). To calculate coverage, the target terrain was divided equally to 400 squares ($10 \times 10$ cm$^2$ for each). Once any of the hybrid robots passed through one of these squares, it was considered covered. They took 10 minutes and 31 seconds to achieve over 80% coverage of the terrain.

**Supplementary Materials**

Fig. S1

Movies S1 to S3

Tables S1 to S2

**Acknowledgments:**

The authors thank Mr. See To Yu Xiang, Dr. Kazuki Kai, Dr Duc Long Le, Mr. Li Rui for their suggestions, and Ms. Kerh Geok Hong, Wendy, for her support and help.

**Funding:** This study was funded by JST (Moonshot R&D Program, Grant Number JPMJMS223A).

**Author contributions:** Conceptualization: Q.L., H.S. Investigation: Q.L., N.V. Methodology: Q.L., K.S., P.T.T.N., G.A.G.N. Visualization: Q.L., K.S. Funding

**Figures**

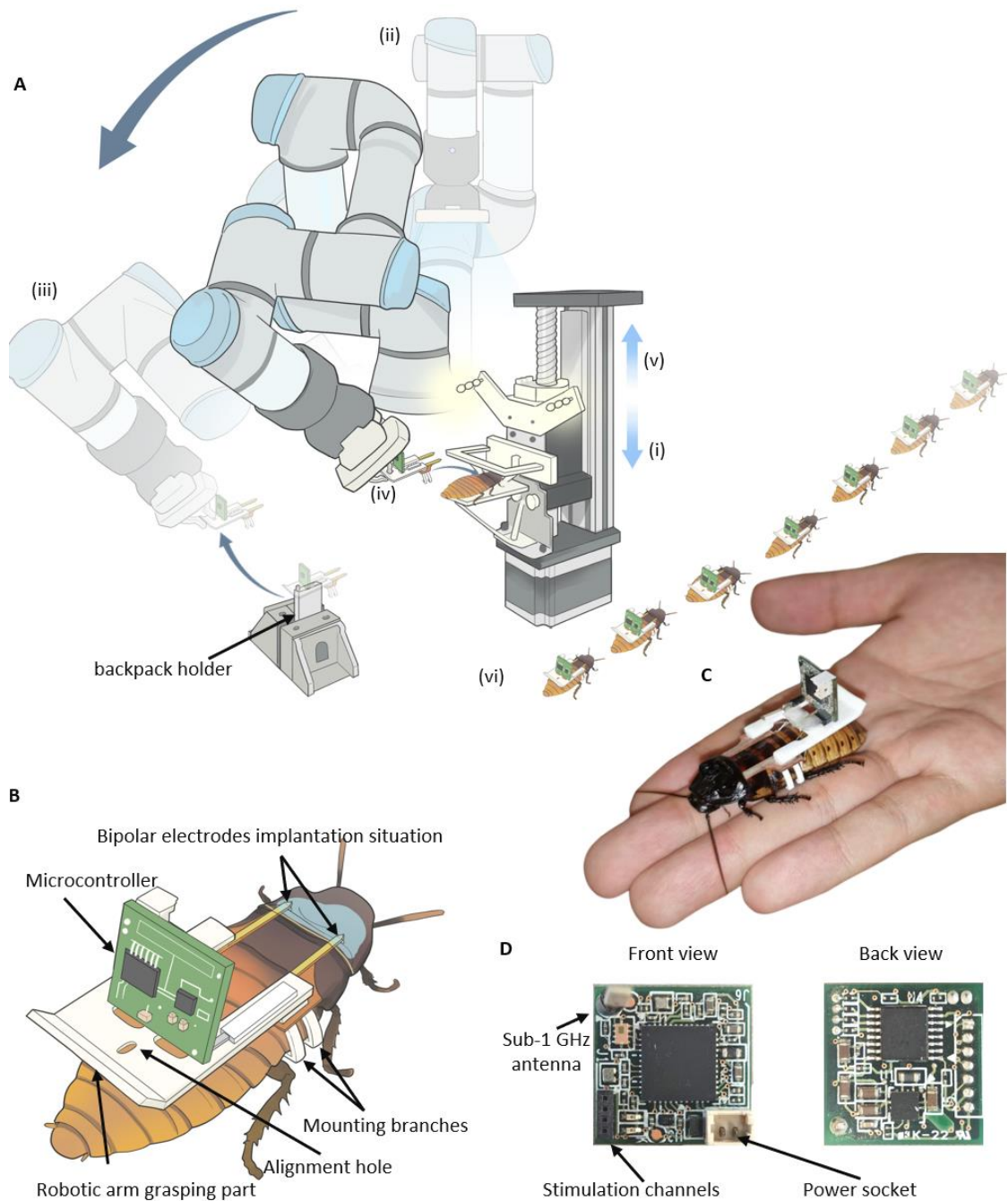

**Fig. 1. Insect-computer hybrid robot and its automatic assembly setup.** (**A**) An anesthetized cockroach was fixed for robotic arm to assemble the backpack. The automatic assembly setup comprised a structure for insect fixation driven by a slide motor, a robotic arm equipped with a gripper for grasping and assembling the backpack, and a depth camera for accurately localizing the insect's body position during assembly.

(**B**) Components of insect-computer hybrid robot. The white mounting structure was sticked with a microcontroller and bipolar electrodes. Grasping part of the backpack could be grasped by robotic arm gripper. The alignment hole was used to place the backpack. Mounting branches hook the cockroach's metathorax. Bipolar electrodes of the backpack implanted the intersegmental membrane between cockroach's pronotum and mesothorax. (**C**) Insect-computer hybrid robot. The cockroach after being assembled with backpack could be controlled for turning and decelerating. (**D**) Microcontroller. Sub-1GHz was used to communicate between insect-computer hybrid robot and workstation. Stimulation signals were sent out from the stimulation channels to the bipolar electrodes to control insect's locomotion. A LiPo battery was connected to the power socket before the hybrid robot was used for the experiment.

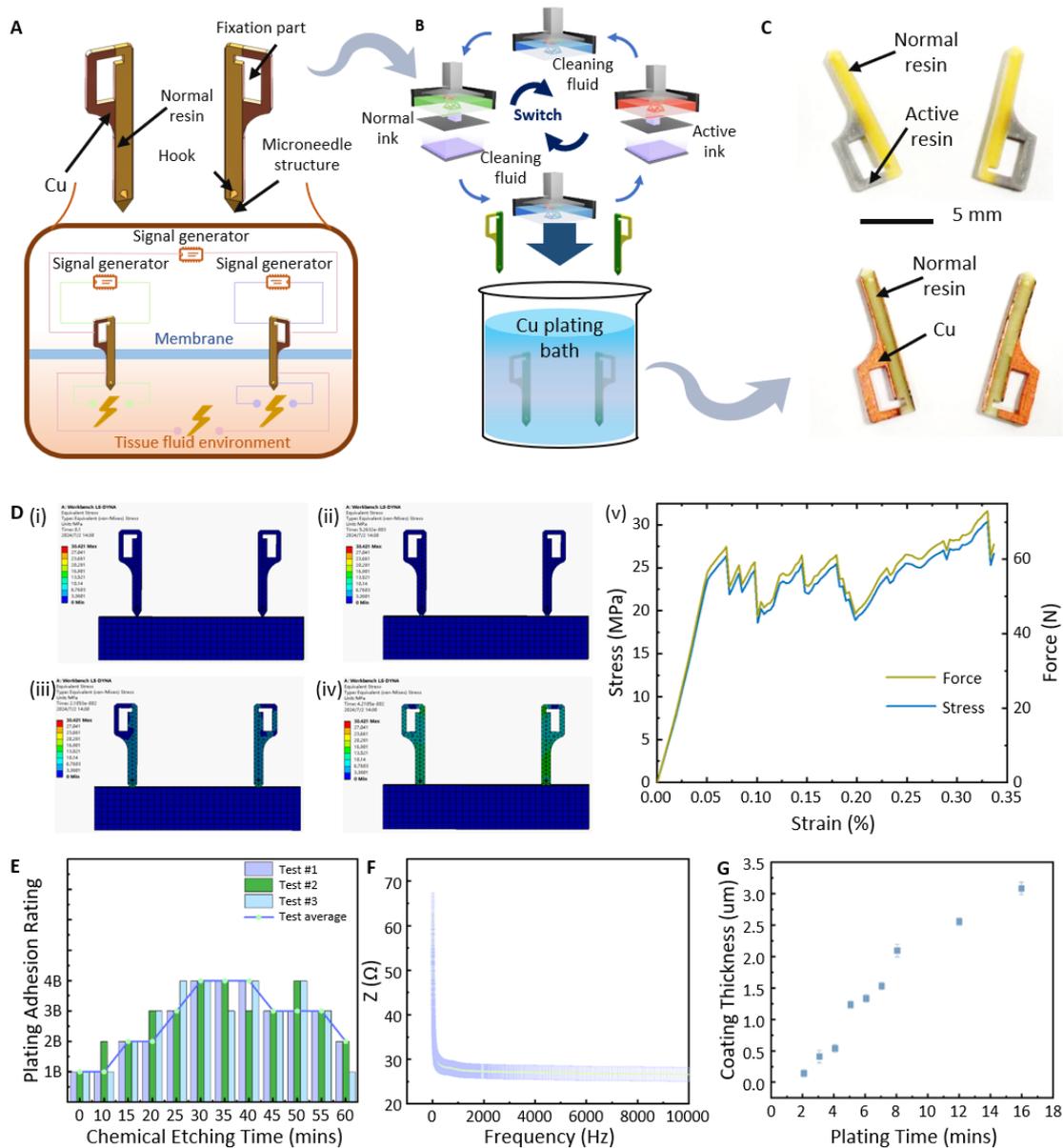

**Fig. 2. Bipolar electrode design, fabrication process and main properties.** (**A**) The custom-designed bipolar electrode utilized a microneedle structure combined with a hook design, enabling rapid membrane penetration and secure self-locking within the punctured membrane. The electrode was composed of normal resin and patterned copper wires to achieve transmission of the stimulation. (**B**) Bipolar electrodes fabricated by multi-material 3D printing technology and subsequent electroless plating process. (**C**) Photos of the bipolar electrodes before and after copper plating. (**D**) Finite element modeling and analysis during the implantation of bipolar electrodes into intersegmental membrane between pronotum and mesothorax of the cockroach. (**E**) Adhesion rating of the metal plating in the bipolar electrode, rated according to the ASTM D3359-09 standard. The introduction of chemical etching ensured the metal adhesion to avoid the degradation of electrical performance due to its detachment during implantation and use. (**F**) Impedance characteristics of bipolar electrodes with conductivity. The impedance consistently measured below 70 Ω, which was significantly lower compared to the impedance values previously recorded with non-invasive electrodes (7). (**G**) Plating thickness as a function of time. The curve illustrated a growth trend, indicating a thicker coating layer with longer plating time.

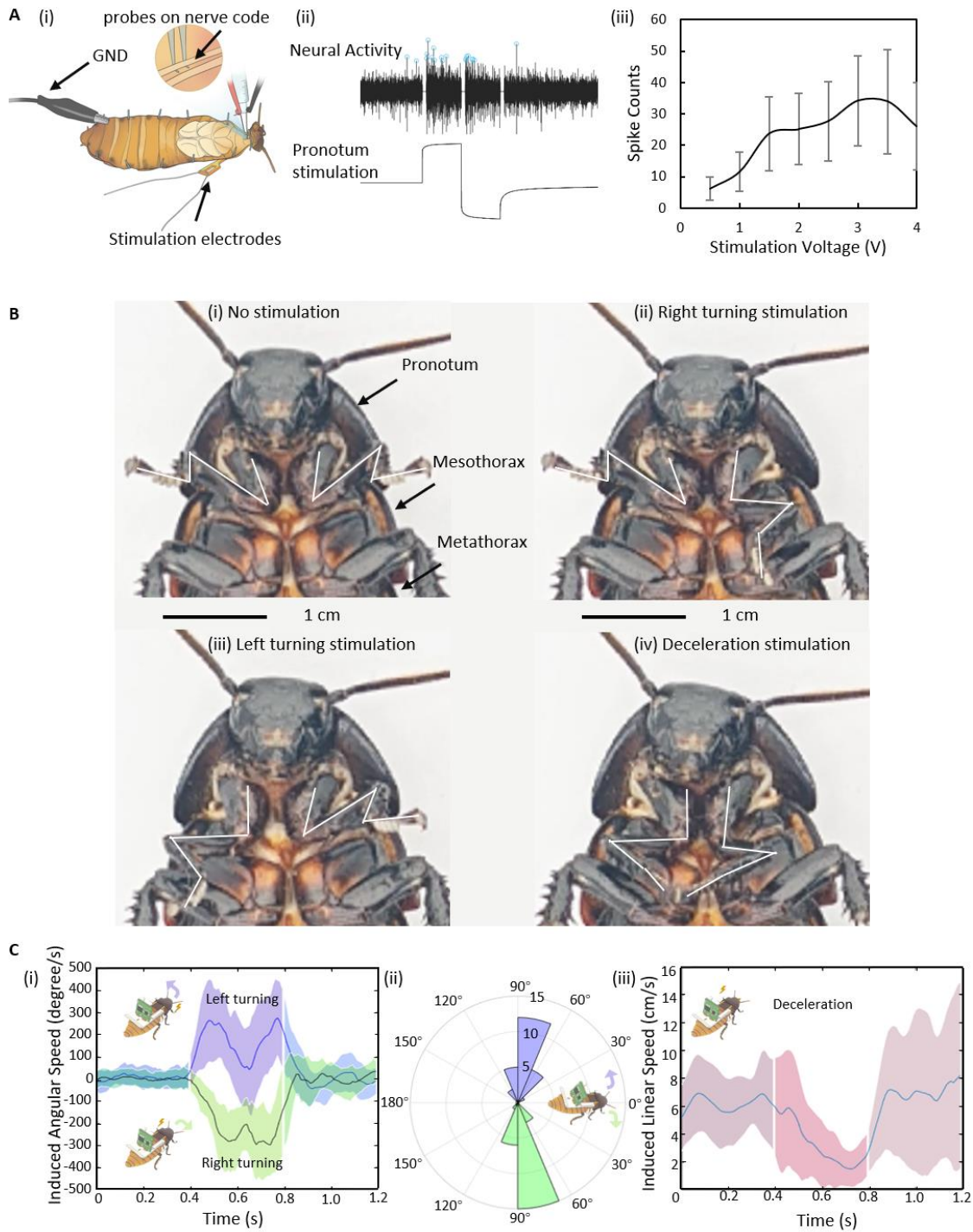

**Fig. 3. Analysis of insect's reaction to electrical stimulation.** (**A**) Insect' neural reaction against the electrical stimulation. i) Neural recording setup. Two probes were attached to the nerve code within the cockroach's neck to capture neural signals. ii) Neural activity and corresponding electrical stimulation. The neural spikes induced by electrical stimulation (marked with blue circles) were quantified to compare responses

under different stimulation voltages. iii) Spike counts at different stimulation voltages (Mean ± SD). (**B**) Insect leg responses to electrical stimulation. i) No stimulation: The insect's forelegs extended. ii) Right-turning stimulation: The insect's left foreleg was stimulated and contracted. iii) Left-turning stimulation: The insect's right foreleg was stimulated and contracted. iv) Deceleration stimulation: Both insect's forelegs were stimulated and contracted. (**C**) Insect locomotion responses to electrical stimulation. i) Induced angular speed during turning stimulation (Mean ± SD). ii) Angular change during turning stimulation (Mean ± SD. iii) Induced linear speed during deceleration stimulation (Mean ± SD).

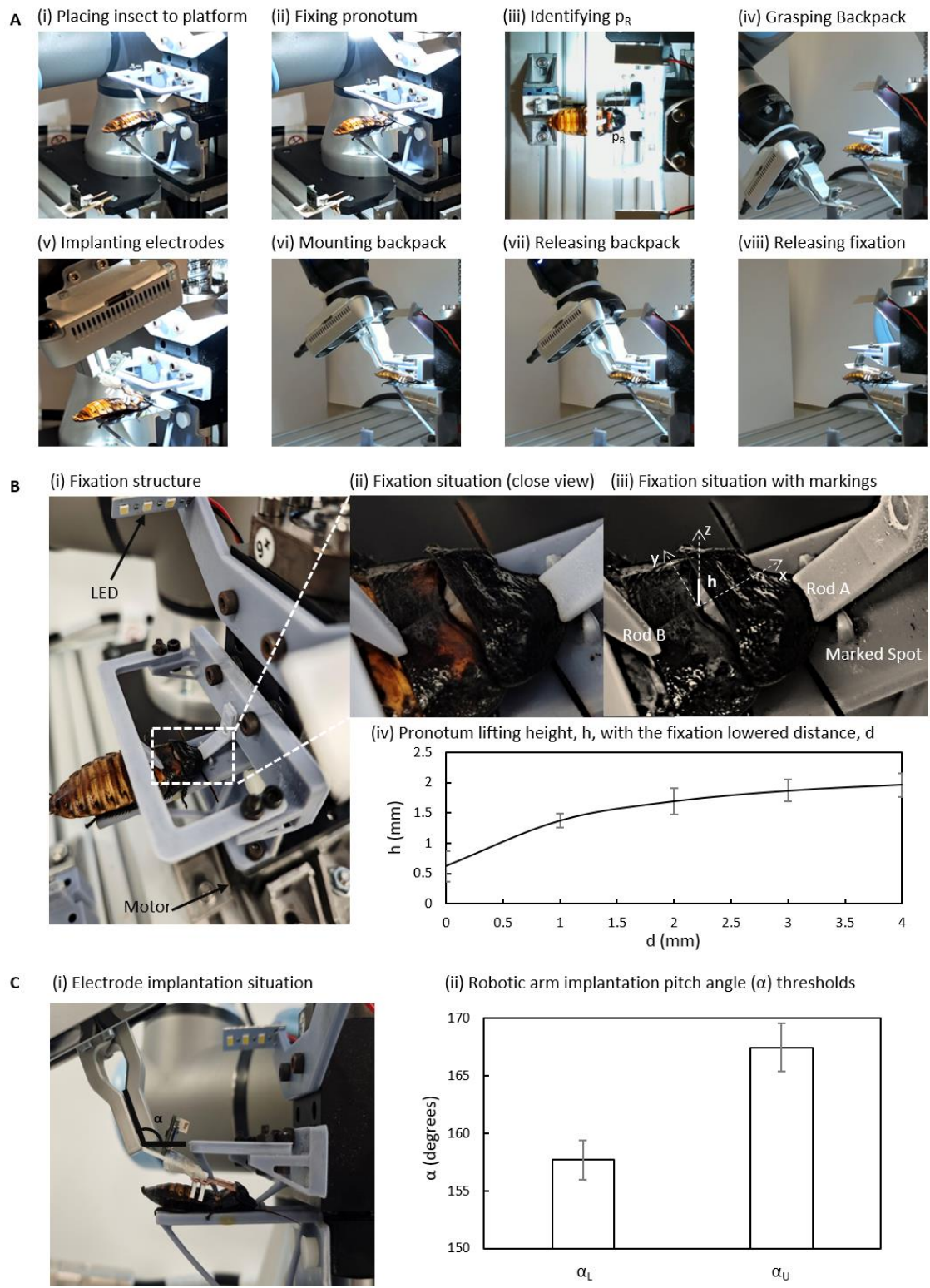

**Fig. 4. Automatic assembly of insect-computer hybrid robots.** (**A**) Detailed process for automatic assembly. i-ii) An anesthetized cockroach on the platform was fixed with a 3D designed structure. iii) The camera captured an image of the cockroach and identified the reference point for bipolar electrodes implantation (green dot). iv-vi)

The robotic arm grasped a backpack, implanted bipolar electrodes, and mounted the backpack. vii-viii) The robotic arm released the backpack and the structure to fix insect retreated. (**B**) Insect fixation. i) Setup of the insect fixation. ii) Close-up view of the fixation of the insect's pronotum. iii) Fixation details with markings. Rod A pressed the anterior pronotum and Rod B pressed the mesothorax, lifting the posterior pronotum. iv) Pronotum lifting height (h) increased with the lowered distance (d) of Rod A. (**C**) Bipolar electrode implantation. i) The implantation pitch angle ($\alpha$) was measured with avoidance of collision with the insect or the structure to fix insect. ii) The lower threshold $\alpha_L$ was measured when the backpack contacted the 3D designed structure, and the upper threshold $\alpha_U$ was determined when the backpack touched the insect. To avoid potential collisions, the midpoint of these thresholds, $\alpha = 162.7$ degrees, was selected.

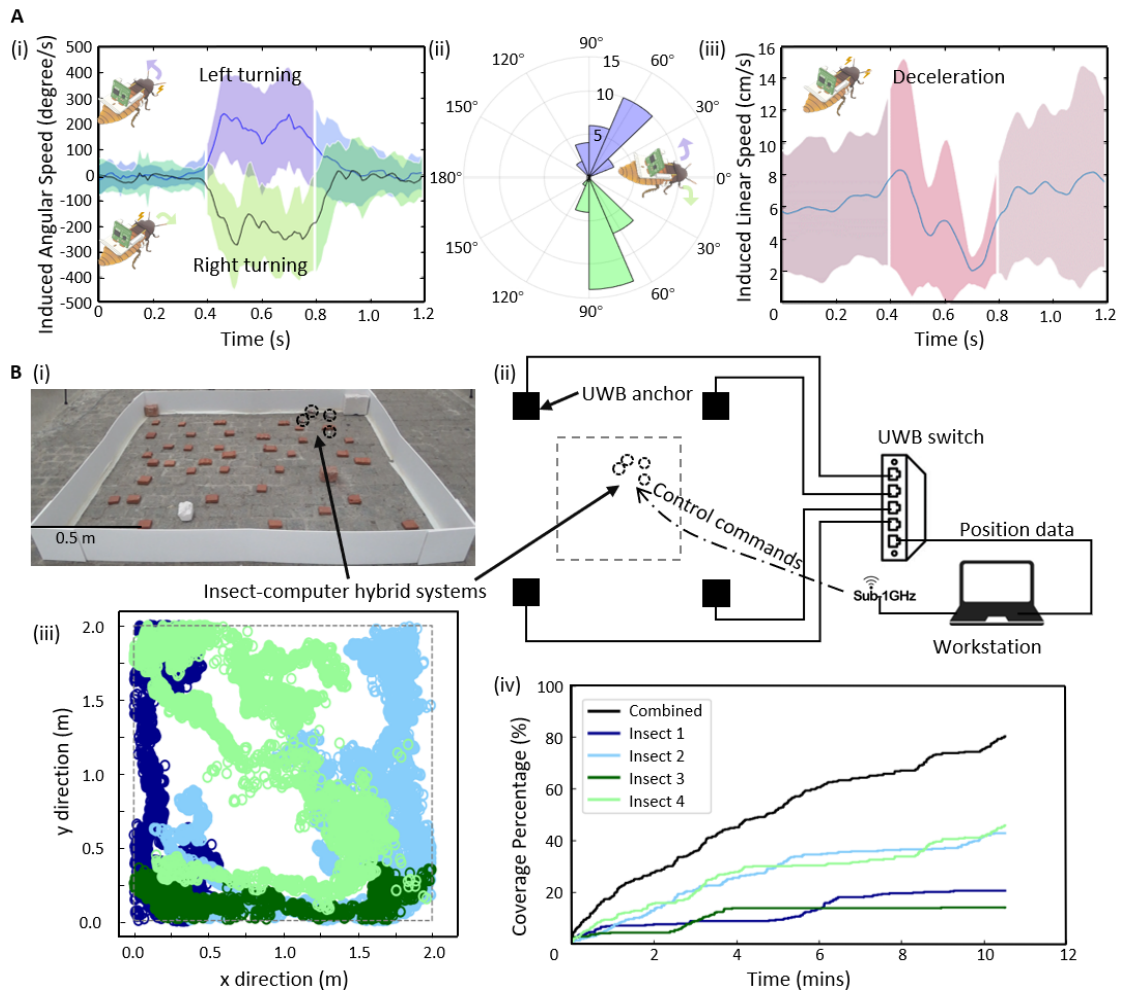

**Fig. 5. Locomotion control and application of automatically assembled insect-computer hybrid robots.** (**A**) Locomotion control of insect-computer hybrid robots. i) Induced angular speed during turning stimulation (Mean ± SD). ii) Angular change during turning stimulation (Mean ± SD). No significant difference in turning angles was observed compared to manually assembled hybrid robots (Student's t-test: $P = 0.62$ for left turns, $P = 0.50$ for right turns). iii) Induced linear speed during the deceleration stimulation (Mean ± SD). No significant difference in decrease of linear speed was observed compared to manually assembled hybrid robots (Student's t-test: $P = 0.21$). (**B**) Covering mission of multiple insect-computer hybrid robots. i) Overview of the obstructed terrain of the cover mission. ii) Setup of the covering experiment. Four UWB anchors were used to track hybrid robots' positions and sent the position data to the workstation for recording. Hybrid robots were controlled by workstation via Sub-1GHz. The target arena was within the gray dashed lines. iii) Insect-computer hybrid robots' trajectories during the mission. iv) Coverage percentage with time. Coverage rate with all four hybrid robots were higher (80.25%) than the single one (from 14.00% to 45.75%), showing the efficiency of the multiple hybrid robots using for the same covering mission.

**Tables**

**Table 1. Comparison of models on the cockroach test samples dataset**

| Model | Params (M) | mIoU | mDSC | MSE ($p_R$) |
|---|---|---|---|---|
| UNet | 31 | 0.8877 | 0.9397 | 2.733 |
| Deeplabv3 | 61 | 0.9058 | 0.9465 | 5.517 |
| TransUNet | 105 | **0.9153** | **0.9550** | **1.707** |
| Segment Anything | 636 | 0.8471 | 0.9161 | 2.387 |

**Table 2: Comparison of loss functions used for TransUNet training**

| Loss Function | mIoU | mDSC | MSE ($p_R$) |
|---|---|---|---|
| BCE Loss | 0.9153 | 0.9550 | 1.707 |
| DSC Loss | **0.9283** | **0.9627** | **1.695** |
| bDoU Loss | 0.9185 | 0.9568 | 1.963 |

**Table 3. Policy parameters of bipolar electrode and implanted structure**

| Object | Density | Young's Modulus | Poisson's Ratio | Yield Strength | Breaking Extensibility |
|---|---|---|---|---|---|
| Bipolar electrode model | 1.26g/cm$^3$ | 330MPa | 0.32 | 24MPa | 0.08 |
| Implanted structure | 1.21g/cm$^3$ | 675MPa | 0.38 | 38MPa | 0.2 |

# Supplementary Materials for

**Title:**

Cyborg Insect Factory: Automatic Assembly System to Build up Insect-computer Hybrid Robot Based on Vision-guided Robotic Arm Manipulation of Custom Bipolar Electrodes

**Authors:**

Qifeng Lin[1], Nghia Vuong[1], Kewei Song[1], Phuoc Thanh Tran-Ngoc[1], Greg Angelo Gonzales Nonato[1], Hirotaka Sato[1*]

**Affiliations:**

[1]School of Mechanical & Aerospace Engineering, Nanyang Technological University; 50 Nanyang Avenue, 639798, Singapore.

*Corresponding author. Email: hirosato@ntu.edu.sg

This file includes:

Fig. S1. Qualitative comparison of segmentation models on the cockroach dataset.

Table S1. Raw mechanical property parameters of the ABS-Like photopolymer used in this study

Table S2. Printing parameters of normal resin and active precursor used in this study

**Other Supplementary Materials for this manuscript include the following:**

Movie S1 (.mp4 format). Demonstration of the front legs movement of the insect-computer hybrid system during stimulation. When the right turning stimulation happens, the left front leg was stimulated to contract and vice versa. When the deceleration stimulation happens, both front legs were stimulated to contract. The contract of front legs shows the insect was under stimulation, indicating that the insect's turning motion was directly due to the stimulation of the associated front leg.

Movie S2 (.mp4 format). Demonstration of the automatic assembly of the insect-computer hybrid system. Insect was firstly fixed by a 3D structure driven by a slide motor. After fixation, its intersegmental membrane between the pronotum and mesothorax was exposed. Gripper of the robotic arm was then activated, and the RealSense Camera scanned the insect to find the specific position of the reference point, $p_R$, on the edge of the pronotum. Afterwards, robotic arm went down to carry the backpack and implanted the bipolar electrodes to the exposed intersegmental membrane. Branches of the backpack were than pressed to fix the backpack to the insect. Finally, robotic arm and slider motor went back to the home position for next assembly. One insect-computer hybrid system was assembled successfully.

Movie S3 (.mp4 format). Demonstration of swarm search using multiple insect-computer hybrid systems. An obstructed outdoor terrain was selected to demonstrate the controllability of the designed hybrid system, especially its outstanding obstacle self-avoidance ability. Four insect-computer hybrid systems were released from a corner of the terrain. UWB localization system tracked the position of each hybrid system. After 10 minutes 31 seconds, the team achieved over 80% coverage.

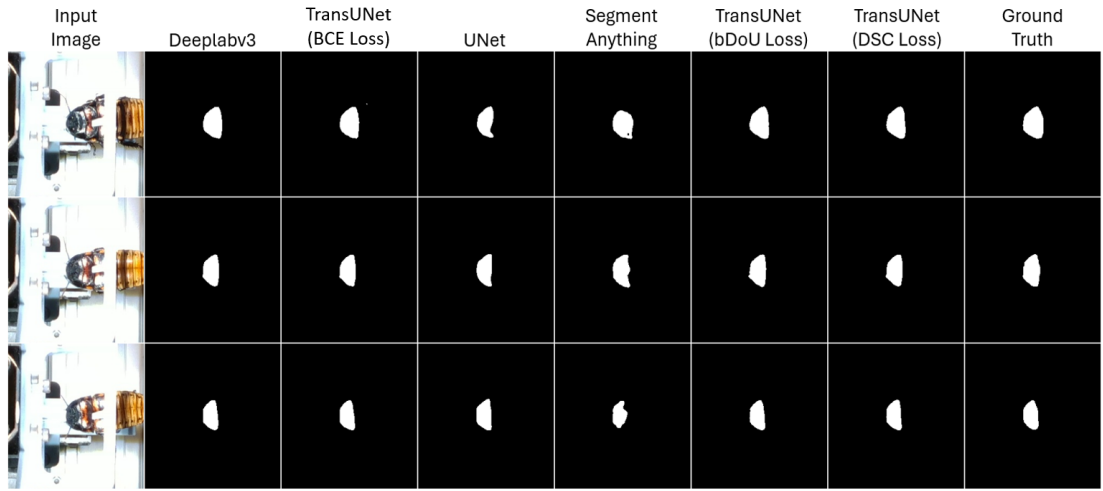

**Fig. S1. Qualitative comparison of segmentation models on the cockroach dataset.** Input images of three cockroaches were compared to select the appropriate model with loss function. Compared to the ground truth segmented pronotum, TransUNet (with BCE Loss) and Deeplabv3 outperformed than the UNet and Segment anything model with more complete outlines. Further comparison on the mIoU, mDSC and MSE ($p_R$) were conducted to show a clear result (Table 1 and Table 2 in the main context).

**Table S1. Raw mechanical property parameters of ABS-like photopolymer used in this study**

| Items | Standard | Value |
|---|---|---|
| Model | / | 603A |
| Attributes | / | Tough resin |
| Appearance | / | multiple colour |
| Viscosity | （cps@25°C） | 260-350 |
| Critical exposure energy | (mJ/cm2) | 8.1~9.0 |
| Tensile Strength (MPa) | ASTM D638 | 46 |
| Tensile modulus (MPa) | ASTM D638 | 2200 |
| Bending strength (MPa) | ASTM D790 | 64 |
| Flexural modulus (MPa) | ASTM D790 | 1900 |
| Elongation (%) | ASTM D638 | 15-25 |
| Notched impact strength (J/m) | ASTM D256 | 40 |
| Heat distortion temperature (°C) | ASTM D 648 @66PSI | 63 |
| Shore hardness (D) | / | 75 |

**Table S2. Printing parameters of normal resin and active precursor used in this study**

| Resin type | Single layer exposure time | Slice thickness |
|---|---|---|
| Normal resin | 6 s - 11 s | 0.05 mm or 0.1 mm |
| Active precursor | 8 s - 12 s | 0.05 mm or 0.1 mm |